\pdfoutput=1
\documentclass[11pt]{article}

\usepackage[]{EMNLP2022}
\usepackage{times}
\usepackage{latexsym}

\usepackage[T1]{fontenc}
\usepackage[utf8]{inputenc}

\usepackage{microtype}

\usepackage{inconsolata}

\usepackage{tabularx} %表格自动换行
\usepackage{array}
\usepackage{multirow}
\usepackage{booktabs}
\usepackage{graphicx} 
\usepackage{url} %网页引用
\usepackage{amsthm,amsmath,amssymb} %希腊字母
\usepackage{mathrsfs} %花体字
\usepackage{bm} %公式中的加粗
\usepackage{pifont} %公式需要
\usepackage{float}
\usepackage{lipsum}
\usepackage{soul}
\usepackage{caption}
\usepackage{subcaption}
\usepackage{enumitem}

\newcommand{\fixme}[1]{}

\newcommand*{\affaddr}[1]{#1} % No op here. Customize it for different styles.
\newcommand*{\affmark}[1][*]{\textsuperscript{#1}}
\newcommand*{\email}[1]{\texttt{#1}}

\title{Improved Universal Sentence Embeddings with Prompt-based Contrastive Learning and Energy-based Learning}

\author{
Yuxin Jiang\affmark[1,2]~~~~Linhan Zhang\affmark[3]~~~~Wei Wang\affmark[1,2]\\
\affaddr{\affmark[1]The Hong Kong University of Science and Technology (Guangzhou)}\\
\affaddr{\affmark[2]The Hong Kong University of Science and Technology}\\
\affaddr{\affmark[3]School of Computer Science and Engineering, The University of New South Wales}\\
\email{yjiangcm@connect.ust.hk},
\email{linhan.zhang@student.unsw.edu.au},
\email{weiwcs@ust.hk}
}

\begin{document}
\maketitle
\begin{abstract}
% due to the distribution variance of inputs between training and evaluation, tuning PLMs with millions of parameters is prone to overfit the training data and result in notoriously poor performance to domain shift. 

% supplement NT-Xent loss when hard negatives are available in supervised datasets. 
% Experiments show that our unsupervised models achieve competitive results compared to previous best methods on standard semantic textual similarity (STS) tasks and state-of-the-art performance on a zero-shot domain transfer STS task; our supervised models achieve new state-of-the-art performance on both standard STS tasks and the zero-shot domain transfer STS task.

Contrastive learning has been demonstrated to be effective in enhancing pre-trained language models (PLMs) to derive superior universal sentence embeddings.
However, existing contrastive methods still have two limitations.
Firstly, previous works may acquire poor performance under domain shift settings, thus hindering the application of sentence representations in practice. We attribute this low performance to the over-parameterization of PLMs with millions of parameters. To alleviate it, we propose PromCSE (Prompt-based Contrastive Learning for Sentence Embeddings), which only trains small-scale \emph{Soft Prompt} (i.e., a set of trainable vectors) while keeping PLMs fixed. %
Secondly, the commonly used NT-Xent loss function of contrastive learning does not fully exploit hard negatives in supervised learning settings. 
To this end, we propose to integrate an Energy-based Hinge loss to enhance the pairwise discriminative power, inspired by the connection between the NT-Xent loss and the Energy-based Learning paradigm. %
Empirical results on seven standard semantic textual similarity (STS) tasks and a domain-shifted STS task both show the effectiveness of our method compared with the current state-of-the-art sentence embedding models.\footnote{Our code is publicly avaliable at \url{https://github.com/YJiangcm/PromCSE}}

\end{abstract}

\section{Introduction}
\label{sec: intro}

Learning universal sentence embeddings \cite{KirosZSZUTF15_skipthought, HillCK16, ConneauKSBB17USE, CerYKHLJCGYTSK18, ReimersG19sbert} which convey high-level semantic information without task-specific fine-tuning is a vital research problem in Natural Language Processing (NLP) communities.
It could benefit a wide range of applications such as information retrieval, question answering, etc \cite{LogeswaranL18}. 
Recently, fine-tuning Pre-trained Language Models (PLMs) \cite{devlin2019bert} with \textit{contrastive learning}, which aims to pull semantically close samples together and push apart dissimilar samples, has achieved extraordinary success in learning universal sentence representations \cite{liu2021mirrorbert, declutr, kim2021sgopt, gao2021simcse, chuang2022diffcse}.
% Contrastive learning aims to pull semantically close samples together and push apart dissimilar samples in the vector space \cite{Chen2020simclr}.
In these works, positive pairs are formed via data augmentation or supervised datasets, whereas negative pairs are derived from different sentences within the same mini-batch. Then contrastive learning objective like normalized temperature-scaled cross-entropy loss (NT-Xent)~\cite{Chen2020simclr, gao2021simcse} is used for optimizing the model parameters. 
% A typical example SimCSE \cite{gao2021simcse} achieves extraordinarily strong performance by utilizing standard dropout as augmentation to construct positive pairs under unsupervised setting while leveraging annotated natural language inference (NLI) datasets under supervised settings.
As a typical example, SimCSE~\cite{gao2021simcse} uses the standard dropout as augmentation for constructing positive pairs and achieves extraordinarily strong performance on seven standard Semantic Textual Similarity (STS) tasks.
% including STS12--STS16~\cite{sts12, sts13, sts14, sts15, sts16}, STS-B~\cite{stsb} and SICK-R~\cite{sickr}.

% Though existing contrastive methods for universal sentence embeddings have shown promising
% results on seven standard Semantic Textual Similarity (STS) tasks, their robustness of domain shift 

\begin{figure}[!t]
	\centering 
	\includegraphics[width=\columnwidth]{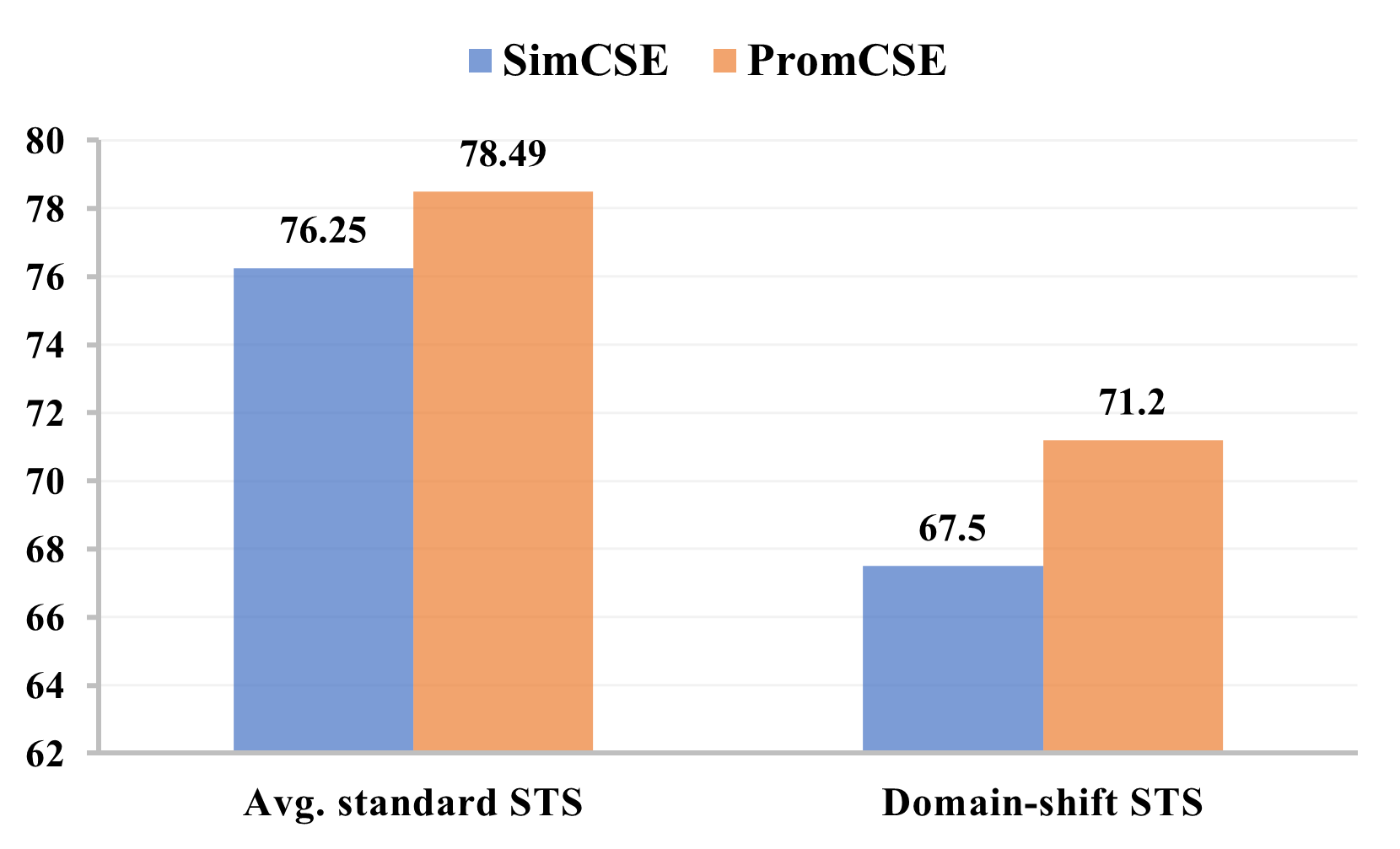}
	\caption{The performance comparison between unsupervised SimCSE and unsupervised PromCSE. Both models are trained on 1 million unlabeled sentences from English Wikipedia.}
	\label{fig: comparison}
\end{figure}

Though effective, existing contrastive methods for learning sentence representations still have two limitations.
\textit{Firstly}, since universal sentence embeddings are often trained on a large corpus and used off-the-shelf on a diverse range of tasks, such domain shifts are commonplace and may pose challenges to the performance. % 
As Figure~\ref{fig: comparison} shows, SimCSE's performance drops significantly when applied on a domain-shifted STS task~\cite{parekh-etal-2021-cococxc}, where the texts are image captions.
Such non-robustness of large PLMs towards domain shifts has also been observed in other studies. \citet{ma2019domain,Lester21sacle} found that tuning PLMs with millions of parameters may result in overfitting to the training data distribution and hence vulnerability to domain shifts. 
% Recently, \cite{Lester21sacle,li2021prefix} proposed prompt tuning, which conditions a frozen PLM with \textit{Soft Prompt} (i.e., a sequence of continuous vectors prepended to the input of PLMs). They showed that prompt tuning is sufficient to be competitive with full model tuning while conferring benefits in robustness to domain shifts. %
\textit{Secondly}, the commonly used NT-Xent loss function in supervised sentence embedding models does not fully exploit the hard negatives.
% provided by Natural Language Inference (NLI) datasets. %
Moreover, recent studies ~\cite{DBLP:conf/cvpr/WangWZJGZL018,DBLP:conf/cvpr/DengGXZ19} have shown that the softmax-based loss is insufficient to acquire the discriminating power.
% In supervised and metric learning literature, it is well-known that hard (i.e., true negative) examples can help guide a learning method to correct its mistakes more quickly~\cite{DBLP:conf/cvpr/SchroffKP15, DBLP:conf/cvpr/SongXJS16}.
Thus, NT-Xent loss in supervised models may not separate positives and hard negatives sufficiently.

In this paper, we propose two techniques to address the above-mentioned limitations. Firstly, we propose the \textbf{Prom}pt-based \textbf{C}ontrastive Learning for \textbf{S}entence \textbf{E}mbeddings (\textbf{PromCSE}) to alleviate the domain shift problem, inspired by prompt tuning~\cite{Lester21sacle,li2021prefix}. Specifically, we modify SimCSE by freezing the entire pre-trained model and add multi-layer learnable \textit{Soft Prompt}, which is simple yet achieves a good balance between the expressiveness and the robustness to distributional changes. 
Secondly, we show that the contrastive learning framework under NT-Xent loss \cite{2020simclr} could be seen as an instance of Energy-Based Learning \cite{DBLP:journals/neco/Hinton02, lecun2006tutorial, DBLP:journals/jmlr/RanzatoBCL07}.
% We find that NT-Xent loss of supervised SimCSE simply regards hard negatives in Natural Language Inference (NLI) datasets as in-batch negatives, which may not fully exploit them.
Inspired by this connection, we propose an Energy-based Hinge (EH) loss to supplement NT-Xent loss under supervised settings, which enhances the pairwise discriminative power by explicitly creating an energy gap between positive pairs and the hardest negative pairs.
% With the supervised Natural Language Inference (NLI) datasets where the "contradictions" are used as "hard negatives", we enhance supervised SimCSE by proposing an Energy-based Hinge (EH) loss to supplement the original contrastive objective.
%
We performed extensive experiments using the seven commonly used STS tasks and another out-of-domain STS task. For the same-domain setting, the unsupervised PromCSE can outperform SimCSE by around 2.2 points and is on par with the current state-of-the-art (SOTA) sentence embedding method on the seven standard STS tasks. For the out-of-domain setting, the proposed unsupervised PromCSE can achieve 3.7 absolute points improvements over SimCSE and even 1.2 absolute points improvements over the current SOTA method, which demonstrates its robustness to domain shifts. Moreover, we also demonstrate that the EH loss can improve supervised SimCSE and PromCSE consistently over multiple pre-trained backbone models, achieving state-of-the-art results among supervised sentence representation learning methods.
% It pushes state-of-the-art results to 82.13\% using BERT$_{base}$ and 85.10\% using RoBERTa$_{large}$.

Our contributions are summarized as follows:
\begin{itemize}[noitemsep]
% (1) To our knowledge, our work is the first to explore and alleviate the domain shift problem of universal sentence embeddings. 
\item We identified two limitations of the SOTA methods for both unsupervised and supervised universal sentence representation learning in their robustness to domain shifts and the formulation of their loss functions. 
\item We propose a multi-layer, prompt-based solution, dubbed PromCSE, as a robust framework for learning sentence embeddings in both the supervised and unsupervised settings. 

\item We proposed the addition of an Energy-based loss function term to the above contrastive learning framework which can further boost the performance of supervised sentence embeddings.

\item Empirical results on seven standard STS tasks and one domain-shifted STS task both verify the effectiveness of our proposed method.
\end{itemize}

\section{Related Work}

\subsection{Sentence Representation Learning}
Learning universal sentence representations has been studied extensively in prior works, roughly categorized into supervised \cite{ConneauK18senteval, CerYKHLJCGYTSK18} and unsupervised approaches \cite{HillCK16, li20bertflow}. Supervised methods train the sentence encoder on datasets with annotations like the supervised Natural Language Inference (NLI) tasks \cite{CerYKHLJCGYTSK18, ReimersG19sbert}. Unsupervised approaches consider deriving sentence embeddings without annotated data, \textit{e.g.}, average GloVe embeddings \cite{glove}, FastSent \cite{HillCK16} and Quick-Thought \cite{LogeswaranL18}. To leverage the rich semantic information implicitly learned by PLMs, recent works have proposed several technics to mitigate the anisotropy issue \cite{DBLP:conf/emnlp/Ethayarajh19, li20bertflow} of PLMs. Post-processing methods like BERT-flow \cite{li20bertflow} and BERT-whitening \cite{su2021whitening} attempt to regularize the semantic space of sentences. Contrastive learning approaches learn sentence embeddings by creating semantically close augmentations and pulling these representations to be closer than representations of random negative examples, which have achieved significant performance improvement \cite{Yan2021consert, liu2021mirrorbert, giorgi-etal-2021-declutr, gao2021simcse, 2022promptbert, shou-etal-2022-amr, zhou-etal-2022-dclr, zhang-etal-2022-arccse, chuang2022diffcse}. 
% One of the most representive work SimCSE \cite{gao2021simcse} uses dropout as data augmentation strategy, which have been shown to be more effective than other complicated augmentation methods. Following SimCSE, recent works such as \cite{2022promptbert, shou-etal-2022-amr, zhou-etal-2022-dclr, zhang-etal-2022-arccse, chuang2022diffcse} have made promising improvements by adding auxiliary objectives, debiasing, etc.

\subsection{Language Model Prompting}
The language model prompting has emerged with the introduction of GPT-3 \cite{gpt3}, which demonstrates promising few-shot performance. Previous works design various discrete prompts manually for specific tasks such as knowledge extraction \cite{DBLP:conf/emnlp/PetroniRRLBWM19}. To
reduce the tedious process of prompt selection, works like \cite{schick2020exploiting, schick2020s, shin2020autoprompt} focus on automatically searching discrete prompts. However, the prompt search over discrete space is time-consuming and sub-optimal due to the continuous nature of neural networks. To solve these issues, \cite{Lester21sacle, li2021prefix, zhong2021factual, liu2021ptuningv2} propose to use soft prompts, which are sets of trainable vectors in the frozen PLMs. These vectors allow the optimization of the downstream tasks in an end-to-end manner. As shown in \cite{Lester21sacle}, PLMs with \textit{Soft Prompts} can often perform better in domain-shift settings.

\subsection{Energy-based Learning}
Energy-based Learning provides a common theoretical framework for many learning models, both probabilistic and non-probabilistic \cite{DBLP:journals/neco/Hinton02, lecun2006tutorial, DBLP:journals/jmlr/RanzatoBCL07}. Energy-Based Models (EBMs) involve four key components: a scalar \textit{energy} function to measure the degree of compatibility between each configuration of the variables; the \textit{inference} algorithm consisting in setting the value of observed variables and finding values of the remaining variables that minimize the energy; the \textit{loss} function which measures the quality of the available energy functions using the training set; the \textit{learning} algorithm consisting in finding an energy function that associates low energies to correct values of the remaining variables, and higher energies to incorrect values. So far, EBMs have been applied in sparse representation learning \cite{DBLP:conf/nips/RanzatoPCL06}, language modeling \cite{DBLP:conf/icml/MnihT12}, text generation \cite{DBLP:conf/iclr/DengBOSR20}, etc.

\section{Methodology}

In this section, we first present \emph{PromCSE}, a prompt-based contrastive learning framework for both unsupervised and supervised sentence representation learning in Section \ref{sec: promcse}.
Then we demonstrate that the contrastive learning framework under NT-Xent loss is an instance of Energy-based Learning in Section~\ref{sec: connect}. Eventually, inspired by Energy-based Learning, we design an Energy-based Hinge loss to supplement NT-Xent loss when hard negatives are available in supervised datasets in Section~\ref{sec: eh loss}.

\subsection{Prompt-based Contrastive Learning}
\label{sec: promcse}
Our prompt-based contrastive learning framework consists of two steps. Firstly, an encoder is built by prepending \emph{Soft Prompt} at \emph{each} layer of the PLM to acquire the sentence representation. Then we optimize the sentence embedding vector space based on the contrastive learning objective.

\paragraph{Sentence Encoder with Soft Prompt}
Fine-tuning is the prevalent way to adapt Transformer-based PLMs as encoders to obtain universal sentence representations. However, model tuning may be over-parameterized and more prone to overfit the training data, to the detriment of similar tasks in different domains.

As an alternative paradigm, prompt tuning~\cite{Lester21sacle, li2021prefix} that conditions a frozen PLM with \textit{Soft Prompt} (i.e., a sequence of continuous vectors prepended to the input of PLMs) has been demonstrated to be competitive with full model tuning while conferring benefits in robustness to domain shifts. By freezing the core language model parameters, prompt tuning prevents the model from modifying its general understanding of language. Instead, prompt representations indirectly modulate the representation of the input. This reduces the model’s ability to overfit a dataset by memorizing specific lexical cues and spurious correlations. Motivated by this, we propose to utilize prompt tuning for universal sentence representations.
During training, we only update the parameters of soft prompts and fix all PLMs parameters. 

Different from \cite{Lester21sacle} which only adds \textit{Soft Prompt} at the input layer, we prepend a sequence of trainable vectors $P^j=\left\{\textbf{p}_{1}^k, ..., \textbf{p}_{l}^k\right\}$ at \emph{each} transformer layer inspired by \cite{liu2021ptuningv2}. Then the $i^{th}$ hidden states at the $j^{th}$ layer $\textbf{h}_{i}^j$ in the Transformers \cite{vaswani2017transformer} are defined as follows:
\begin{equation}
\textbf{h}_{i}^j=\left\{
\begin{array}{ll}
\textbf{e}_{i}^j,& \text{$ j = 0 \wedge i > k $ } \\
\textbf{p}_{i}^j,& \text{$ i \leq k $ } \\
Trans(\textbf{h}^{j-1})_{i},& \text{otherwise}
\end{array}
\right.
\end{equation}
where $Trans()$ denotes the forward function of the Transformer block layer and $\textbf{e}_{i}$ denotes the fixed token embedding vector at the input layer. Compared with \cite{Lester21sacle}, this enables gradients to be backward updated at each layer and better complete the learning tasks. 
During the training, sentences are fed into the frozen PLM with the prepended \textit{Soft Prompt}, and we add an MLP layer over the $[CLS]$ hidden state from the last layer of PLM to obtain the sentence embeddings. 

\paragraph{Contrastive Learning Objective}%
We use the most widely adopted training objective NT-Xent loss~\cite{Chen2020simclr, gao2021simcse}, which has been applied in previous sentence and image representation learning methods. Given a set of paired sentences $\mathcal{D}=\left\{ (X_{i}, X_{i}^+) \right\}_{i=1}^m$ where $X_{i}$ and $X_{i}^+$ are semantically close, we regard $X_{i}^+$ as positive of $X_{i}$ and other sentences in the same mini-batch as negatives. Let $\textbf{h}_{i}$ and $\textbf{h}_{i}^+$ denote the sentence embeddings of $X_{i}$ and $X_{i}^+$, then NT-Xent loss for a single sample in a mini-batch of size $N$ can be formulated as follows:
\begin{equation}
\mathcal{L}_{CL}=-\log\frac{e^{sim(\textbf{h}_{i}, \textbf{h}_{i}^+)/\tau}}{\sum_{j=1}^Ne^{sim(\textbf{h}_{i}, \textbf{h}_{j}^+)/\tau}}
\label{equation: contrastive}
\end{equation}
where $\tau$ is a temperature hyperparameter and $sim(\textbf{h}_{1}, \textbf{h}_{2})$ is the cosine similarity function.

We follow SimCSE~\cite{gao2021simcse} that constructs positive pairs by feeding the same sentence to the sentence encoder twice with diverse dropout masks when only unlabeled text data is available.

\subsection{Connecting Contrastive Learning with Energy-based Learning}
\label{sec: connect}
Given a set of training samples $\mathcal{S}=\left\{(X_i, Y_i), i=1\ldots N\right\}$ where $X$ and $Y$ are two variables, Energy-Based Models (EBMs) use an scalar \textit{energy function} $E(W,Y_i,X_i)$ indexed by parameter $W$ to measure the compatibility between two variables. Note that small energy values correspond to highly compatible configurations of the variables, while large energy values correspond to highly incompatible configurations.
The generalized negative log-likelihood loss of EBMs \cite{lecun2006tutorial}, which stems from a probabilistic formulation of the learning problem in terms of the maximum conditional probability principle, is defined as follows:
\begin{equation}
\mathcal{L}_{nll}=E(W,Y_i,X_i)+\mathcal{F}_{\beta}(W, \mathcal{Y}, X_i)
\end{equation}
where $\mathcal{Y}$ is the set of all possible values of $Y$, $\mathcal{F}$ is the \textit{free energy} of the ensemble $\left\{E(W,y,X_i), y\in \mathcal{Y}\right\}$:
\begin{equation}
\mathcal{F}_{\beta}(W, \mathcal{Y}, X_i)=
\frac{1}{\beta}\log\left(\int_{y\in \mathcal{Y}}e^{-\beta E(W,y,X_i)}\right)
\end{equation}
where $\beta$ is a positive constant akin to an inverse temperature. Consequently, 
\begin{equation}
\mathcal{L}_{nll}\propto
-\log\frac{e^{-\beta \text{$E(W,Y_i,X_i)$}}}{\int_{y\in \mathcal{Y}}e^{-\beta E(W,y,X_i)}}
\label{equation: nll}
\end{equation}

Considering $X_i$ and $Y_i$ are both sentences under the \textit{implicit constraint} that $X_i$ and $Y_i$ are positive pairs, we can define the energy function $E$ as
\begin{equation}
E(W,Y_i,X_i) = -sim(f(X_i),f(Y_i))
\label{equation: energy}
\end{equation}
where $f$ is the sentence encoder parameterized by $W$. According to Equation (\ref{equation: energy}), the loss in Equation (\ref{equation: nll}) can be rewritten as
\begin{equation}
\mathcal{L}_{nll}\propto
-\log\frac{e^{sim(f(X_i),f(Y_i))/ \frac{1}{\beta}}}
{\int_{y\in \mathcal{Y}}e^{sim(f(X_i),f(y))/ \frac{1}{\beta}}}
\end{equation}
Therefore, we can see that the contrastive loss in Equation~\eqref{equation: contrastive} can be deemed as a special case of the Energy-based negative log-likelihood loss.

%%!! then? lack of connection to the rest

\subsection{Energy-based Hinge Loss}
\label{sec: eh loss}
NLI datasets \cite{snli, mnli} that contain entailment, neutral, and contradiction sentence pairs have shown great success in supervised sentence embedding learning \cite{ConneauKSBB17USE, ReimersG19sbert}.
Supervised SimCSE incorporate annotated sentence pairs in contrastive learning by leveraging entailment pairs as positives and extending in-batch negatives with contradiction pairs, namely \textit{hard negatives}.
% Supervised SimCSE leverages entailment pairs as positives for contrastive learning and demonstrates that extending in-batch negatives with contradiction pairs, namely \textit{hard negatives}, can further improve the performance. 
The NT-Xent loss for supervised SimCSE is:
\begin{equation}
\mathcal{L}_{CL}=-\log\frac{e^{sim(\textbf{h}_{i}, \textbf{h}_{i}^+)/\tau}}{\sum_{j=1}^N(e^{sim(\textbf{h}_{i}, \textbf{h}_{j}^+)/\tau}+e^{sim(\textbf{h}_{i}, \textbf{h}_{j}^-)/\tau})}
\label{equation: contrastive_sup}
\end{equation}
where $\textbf{h}_{i}, \textbf{h}_{i}^+, \textbf{h}_{j}^-$ correspond to the embeddings of premise, entailment hypotheses and contradiction hypotheses. 
Compared with in-batch negatives, hard negatives are more syntactically similar to the anchor, thus making them more likely to be misidentified as positives by the model.
In supervised and metric learning literature, it is well-known that hard (i.e., true negative) examples can help guide a learning method to correct its mistakes more quickly~\cite{DBLP:conf/cvpr/SchroffKP15, DBLP:conf/cvpr/SongXJS16}.
However, the softmax-based NT-Xent loss is shown to be insufficient to acquire the discriminating power~\cite{DBLP:conf/cvpr/WangWZJGZL018,DBLP:conf/cvpr/DengGXZ19}, which may not adequately separate hard negatives and positives.
% However, \cite{WangL21a_understandcl} show that though NT-Xent loss is a hardness-aware loss function, it could still underperform a simpler contrastive loss with hard negatives.
% Since contradiction hypotheses are more syntactically similar to premises compared with in-batch negatives, thus benefiting efficient learning more, Equation (\ref{equation: contrastive_sup}) may not make full exploitation of them.
%%!! use simpler sentences (for important things)
%
Besides, when the temperature $\tau \rightarrow 0^+$, NT-Xent loss degenerates to a triplet loss with a margin of 0 \cite{WangL21a_understandcl}. The small $\tau=0.05$ used in SimCSE avoids this case but may still cause the sentence representations insufficiently discriminative and, as a result, not sufficiently robust to noise due to the small margin.  

To alleviate the above-mentioned problem and inspired by the Energy-based Learning \cite{lecun2006tutorial}, we propose to use the Energy-based Hinge (EH) loss to supplement the original contrastive objective. We first give the following definition:
\paragraph{Definition 1} \textit{Suppose $Y$ is a discrete variable. For a training sample $(X_i, Y_i)$, the \textbf{most offending incorrect answer} $\hat{Y}_i$ is the one that has the lowest energy among all answers that are incorrect:}
\begin{equation}
\hat{Y}_i=\mathop{\arg\min}\limits_{Y\in\mathcal{Y} \land Y \neq Y_i}
E(W,Y,X_i)
\end{equation}
Accordingly, the Energy-based Hinge (EH) loss is defined as follows:
\begin{equation}
[m + E(W,Y_i,X_i) - E(W,\hat{Y}_i,X_i)]_{+}
\label{equation: eh}
\end{equation}
where $m \ge 0$ is the margin, and $[x]_{+}\equiv\max(0,x)$. Combining Equation (\ref{equation: energy}) with Equation (\ref{equation: eh}), we can derive the energy-based hinge loss for sentence embeddings:
\begin{equation}
\mathcal{L}_{EH}=
[m + sim(\textbf{h}_{i}, \hat{\textbf{h}}_{i}) - sim(\textbf{h}_{i}, \textbf{h}_{i}^+)]_{+}
% \begin{aligned}
% [m + \max(\max\limits_{i\neq j} sim(\textbf{h}_{i}, \textbf{h}_{j}^+), \max sim(\textbf{h}_{i}, \textbf{h}_{j}^-)) \\
% - sim(\textbf{h}_{i}, \textbf{h}_{i}^+)]_{+}
% \end{aligned}
\label{equation: eh_1}
\end{equation}
The EH loss enhances the pairwise discriminative power by maximizing the decision margin $m$ in the semantic space.
During the training, we use the nearest sample among in-batch negatives and hard negatives
% sentence among in-batch negatives as well as hard negatives which acquires the highest similarity with the anchor 
to approximate the \textit{most offending incorrect answer}; this works empirically well as we observed that it is often the corresponding contradiction hypothesis. 
% We do not directly use the corresponding contradiction hypothesis as the \textit{most offending incorrect answer}, since human annotation may cause some bias. 
% \begin{equation}
% \begin{split}
% & sim(\textbf{h}_{i}, \hat{\textbf{h}}_{i})= \\
% & \max(\max\limits_{j \in [1,N], j\neq i} sim(\textbf{h}_{i}, \textbf{h}_{j}^+), \max\limits_{j \in [1,N]} sim(\textbf{h}_{i}, \textbf{h}_{j}^-))
% \end{split}
% \end{equation}
Eventually, we can enhance the optimization objective for our \textbf{supervised} models with the combination of Equation (\ref{equation: contrastive_sup}) and (\ref{equation: eh_1})
\begin{equation}
\mathcal{L}=\mathcal{L}_{CL} + \lambda \cdot \mathcal{L}_{EH}
\label{equation: total_loss}
\end{equation}
where $\lambda$ is a weighting coefficient. We set $\lambda$ to 10 empirically because the scale of $\mathcal{L}_{EH}$ is around ten smaller than $\mathcal{L}_{CL}$ during training.

\section{Experiments}
Our experiments are composed of two parts. We first verify the effectiveness of our proposed approach on seven standard STS tasks in Section \ref{sec: standard sts}. Then we evaluate the domain shift robustness of our approach by testing on a domain shift STS task in Section \ref{sec: transfer sts}. 

\begin{table*}[!t]
\small
\centering
\begin{tabular}{lcccccccc}
\toprule
\textbf{Model}                      & \textbf{STS12} & \textbf{STS13} & \textbf{STS14} & \textbf{STS15} & \textbf{STS16} & \textbf{STS-B} & \textbf{SICK-R} & \textbf{Avg.}  \\
\midrule
\midrule
\multicolumn{9}{c}{\textit{Unsupervised models}} \\\midrule
GloVe embeddings (avg.)$^\clubsuit$    & 55.14 & 70.66 & 59.73 & 68.25 & 63.66 & 58.02 & 53.76  & 61.32 \\
BERT$_{base}$ (first-last avg.)$^\diamondsuit$ & 39.70  & 59.38 & 49.67 & 66.03 & 66.19 & 53.87 & 62.06  & 56.70  \\
BERT$_{base}$-flow$^\diamondsuit$              & 58.40  & 67.10  & 60.85 & 75.16 & 71.22 & 68.66 & 64.47  & 66.55 \\
BERT$_{base}$-whitening$^\diamondsuit$         & 57.83 & 66.90  & 60.9  & 75.08 & 71.31 & 68.24 & 63.73  & 66.28 \\
IS-BERT$_{base}$$^\heartsuit$                & 56.77 & 69.24 & 61.21 & 75.23 & 70.16 & 69.21 & 64.25  & 66.58 \\
CT-BERT$_{base}$$^\diamondsuit$                & 61.63 & 76.80  & 68.47 & 77.50  & 76.48 & 74.31 & 69.19  & 72.05 \\
ConSERT$_{base}$$^\spadesuit$                & 64.64 & 78.49 & 69.07 & 79.72 & 75.95 & 73.97 & 67.31  & 72.74 \\
Mirror-BERT$_{base}$$^\dag$            & 67.40  & 79.60  & 71.30  & 81.40  & 74.30  & 76.40  & 70.30   & 74.40  \\
SG-OPT-BERT$_{base}$$^\ddag$            & 66.84 & 80.13 & 71.23 & 81.56 & 77.17 & 77.23 & 68.16  & 74.62 \\
SimCSE-BERT$_{base}$$^\diamondsuit$            & 68.40  & 82.41 & 74.38 & 80.91 & 78.56 & 76.85 & \textbf{72.23}  & 76.25 \\
% ESimCSE-BERT$_{base}$$^\blacklozenge$              & 73.40 & 83.27 & 77.25 & 82.66 & 78.81 & 80.17 & 72.30  & 78.27 \\
% DCLR-BERT$_{base}$$^\blacklozenge$              & 70.81 & 83.73 & 75.11 & 82.56 & 78.44 & 78.31 & 71.59  & 77.22 \\
% ArcCSE-BERT$_{base}$$^\vartriangle$            & 72.08 & 84.27 & 76.25 & 82.32 & 79.54 & 79.92 & \textbf{72.39}  & 78.11 \\
DiffCSE-BERT$_{base}$$^\blacklozenge$           & \underline{72.28} & 84.43 & 76.47 & 83.90  & \underline{80.54} & 80.59 & \underline{71.23}  & \underline{78.49} \\
PromptBERT$_{base}$$^\vartriangle$             & 71.56 & \underline{84.58} & \textbf{76.98} & \textbf{84.47} & \textbf{80.60}  & \textbf{81.60}  & 69.87  & \textbf{78.54} \\
$*$ PromCSE-BERT$_{base}$         & \textbf{73.03} & \textbf{85.18} & \underline{76.70}  & \underline{84.19} & 79.69 & \underline{80.62} & 70.00     & \underline{78.49} \\
\midrule
\midrule
\multicolumn{9}{c}{\textit{Supervised models}} \\\midrule
InferSent-GloVe$^\clubsuit$                      & 52.86 & 66.75 & 62.15 & 72.77 & 66.87 & 68.03 & 65.65  & 65.01 \\
Universal Sentence Encoder$^\clubsuit$           & 64.49 & 67.80  & 64.61 & 76.83 & 73.18 & 74.92 & 76.69  & 71.22 \\
SBERT$_{base}$$^\clubsuit$                            & 70.97 & 76.53 & 73.19 & 79.09 & 74.30  & 77.03 & 72.91  & 74.89 \\
SBERT$_{base}$-flow$^\diamondsuit$                       & 69.78 & 77.27 & 74.35 & 82.01 & 77.46 & 79.12 & 76.21  & 76.60  \\
SBERT$_{base}$-whitening$^\diamondsuit$                  & 69.65 & 77.57 & 74.66 & 82.27 & 78.39 & 79.52 & 76.91  & 77.00    \\
CT-SBERT$_{base}$$^\diamondsuit$                         & 74.84 & 83.20  & 78.07 & 83.84 & 77.93 & 81.46 & 76.42  & 79.39 \\
ConSERT-BERT$_{base}$$^\spadesuit$                     & 74.07 & 83.93 & 77.05 & 83.66 & 78.76 & 81.36 & 76.77  & 79.37 \\
SimCSE-BERT$_{base}$$^\diamondsuit$                      & 75.30  & 84.67 & 80.19 & 85.40  & 80.82 & 84.25 & 80.39  & 81.57 \\
$*$ SimCSE-BERT$_{base}$ (reproduce)        & 75.13 & 84.35 & 80.26 & 85.45 & 80.83 & 84.29 & 80.39  & 81.53 \\
$*$ SimCSE-BERT$_{base}$ + EH      & 75.22 & \underline{84.93} & \textbf{81.37} & \underline{85.94} & 80.94 & \textbf{84.78} & 80.38  & \underline{81.94} \\
$*$ PromCSE-BERT$_{base}$                   & \underline{75.58} & 84.33 & 79.67 & 85.79 & \underline{81.24} & 84.25 & \underline{80.79}  & 81.81 \\
$*$ PromCSE-BERT$_{base}$ + EH     & \textbf{75.96} & \textbf{84.99} & \underline{80.44} & \textbf{86.83} & \textbf{81.30}  & \underline{84.40}  & \textbf{80.96}  & \textbf{82.13} \\
\midrule
SimCSE-RoBERTa$_{base}$$^\diamondsuit$                   & 76.53 & 85.21 & 80.95 & 86.03 & 82.57 & 85.83 & \underline{80.50}   & 82.52 \\
$*$ SimCSE-RoBERTa$_{base}$ + EH   & 76.83 & 85.67  & \underline{81.57} & 86.35 & 82.72 & 86.84 & \textbf{80.56}  & 82.86 \\
$*$ PromCSE-RoBERTa$_{base}$                & \underline{76.75} & \underline{85.86} & 80.98 & \underline{86.51} & \underline{83.51} & \textbf{86.58} & 80.41  & \underline{82.94} \\
$*$ PromCSE-RoBERTa$_{base}$ + EH  & \textbf{77.51} & \textbf{86.15} & \textbf{81.59} & \textbf{86.92} & \textbf{83.81} & \underline{86.35} & 80.49  & \textbf{83.26} \\
\midrule
SimCSE-RoBERTa$_{large}$$^\diamondsuit$                  & 77.46 & 87.27 & 82.36 & 86.66 & 83.93 & 86.70  & 81.95  & 83.76 \\
$*$ SimCSE-RoBERTa$_{large}$ + EH  & 78.01 & 87.65 & 82.55 & 87.21 & 84.19 & 86.95 & 82.03  & 84.08 \\
$*$ PromCSE-RoBERTa$_{large}$               & \underline{79.14} & \underline{88.64} & \underline{83.73} & \underline{87.33} & \underline{84.57} & \underline{87.84} & \underline{82.07}  & \underline{84.76} \\
$*$ PromCSE-RoBERTa$_{large}$ + EH & \textbf{79.56} & \textbf{88.97} & \textbf{83.81} & \textbf{88.08} & \textbf{84.96} & \textbf{87.87} & \textbf{82.43}  & \textbf{85.10} \\
\bottomrule
\end{tabular}
\caption{\label{tab: standard sts}
The performance of different sentence embedding models on test sets of STS tasks (Spearman’s correlation). The best performance and the second-best performance methods are denoted in bold and underlined fonts respectively. $\clubsuit$: results from \cite{ReimersG19sbert}; $\diamondsuit$: results from \cite{gao2021simcse}; $\heartsuit$: results from \cite{zhang-etal-2020-isbert}; $\spadesuit$: results from \cite{Yan2021consert}; $\dag$: results from \cite{liu2021mirrorbert}; $\ddag$: results from \cite{kim2021sgopt}; $\blacklozenge$: results from \cite{chuang2022diffcse}; $\vartriangle$: results from \cite{2022promptbert}; $*$ : results from our experiments; + EH: adding the Energy-based Hinge loss as shown in Equation (\ref{equation: total_loss}).
}
\end{table*}

\subsection{Standard STS}
\label{sec: standard sts}
\subsubsection{Setups}
\paragraph{Dataset and Metric}
We use seven standard STS datasets including STS tasks 2012-2016 \cite{sts12, sts13, sts14, sts15, sts16}, STS Benchmark \cite{stsb} and SICK-Relatedness \cite{sickr} for our experiments. Texts of these datasets are from news, forums, lexical definitions, etc. Each sample in these datasets contains a pair of sentences as well as a semantic similarity score ranging from 0 to 5. We use SentEval toolkit \cite{ConneauK18senteval} for evaluation and report the Spearman’s correlation on test sets following previous works \cite{ReimersG19sbert, gao2021simcse}.
\paragraph{Baselines}
We compare unsupervised and supervised PromCSE to previous state-of-the-art sentence embedding methods. Unsupervised baselines comprise average GloVe embeddings \cite{glove}, average BERT embeddings \cite{gao2021simcse}, and post-processing methods such as BERT-flow \cite{li20bertflow} and BERT-whitening \cite{su2021whitening}. We also introduce strong unsupervised baselines using contrastive learning, including IS-BERT \cite{zhang-etal-2020-isbert}, CT-BERT \cite{carlsson2020ctbert}, ConSERT \cite{Yan2021consert}, Mirror-BERT \cite{liu2021mirrorbert}, SG-OPT \cite{kim2021sgopt}, SimCSE \cite{gao2021simcse}, DiffCSE \cite{chuang2022diffcse} and PromptBERT \cite{2022promptbert}. 
Methods taking extra supervision include InferSent \cite{ConneauKSBB17USE}, Universal Sentence Encoder \cite{CerYKHLJCGYTSK18}, SBERT \cite{ReimersG19sbert} along with applying BERT-flow, whitening and CT on it, ConSERT \cite{Yan2021consert} and SimCSE \cite{gao2021simcse}.
\paragraph{Implementation Details}
We implement our models based on Huggingface’s transformers \cite{huggingface}, where we also obtain the pre-trained checkpoints of BERT \cite{devlin2019bert} and RoBERTa \cite{liu2019roberta}.
We use the identical training data as SimCSE \cite{gao2021simcse}. Specifically, we train unsupervised PromCSE on 1 million randomly sampled sentences from English Wikipedia for one epoch, and train supervised PromCSE on the combination of MNLI \cite{mnli} and SNLI \cite{snli} datasets for ten epochs.
% We evaluate our model every 125 training steps on the development set of STS-B, and the best checkpoint is used for the final evaluation on test sets.
The training proceeds with the default random seed 42 for one run, the same as SimCSE. The training details of hyperparameters are shown in Appendix \ref{sec:appendix a}.

\subsubsection{Results}
Table \ref{tab: standard sts} shows that our unsupervised PromCSE-BERT$_{base}$ significantly outperforms SimCSE-BERT$_{base}$ and raises the averaged Spearman’s correlation from 76.25\% to 78.49\%. Besides, it can acquire competitive results with current state-of-the-art DiffCSE-BERT$_{base}$ and PromptBERT$_{base}$. Note that although PromptBERT applies prompting to contrastive learning, it requires fine-tuning the whole PLM and manually designing discrete prompts \cite{2022promptbert}. 
Using supervised NLI datasets, PromCSE also surpasses SimCSE consistently based on various PLMs.
Incorporating the Energy-based Hinge loss under supervised settings can further enhance SimCSE as well as PromCSE consistently over multiple pre-trained backbone models. It pushes state-of-the-art results to 82.13\% using BERT$_{base}$ and 85.10\% using RoBERTa$_{large}$.

\subsection{Domain-Shifted STS}
\label{sec: transfer sts}
\subsubsection{Setups}
\paragraph{Dataset and Metric}
The cumbersome data annotation leads to few datasets for STS tasks. Fortunately, we find a dataset with a different domain from the training corpus and the standard STS tasks. 
Crisscrossed Captions (CxC)~\cite{parekh-etal-2021-cococxc} extends the English MS-COCO \cite{mscoco} 5k dev and test sets with continuous (0-5) human similarity annotations, and it supports evaluation for correlation measures that compare model rankings with rankings derived from human similarity judgments for text-text comparisons.
We use the STS task of CxC, whose texts are all image captions, to evaluate the domain-shifted robustness of various sentence embedding models.

Due to CxC’s dense annotation where the scores between many pairs are themselves correlated, we choose a sampled Spearman's bootstrap correlation as the evaluation metric following~\cite{parekh-etal-2021-cococxc}. For each correlation estimate, we sample half of the queries and for each selected query, we choose one of the items for which CxC supplies a paired rating. We compute Spearman’s $r$ between the CxC scores and the model scores for the selected pairs. The final correlation is the average over 1000 of these bootstrap samples.

\paragraph{Baselines}
We compare our unsupervised and supervised models to current SOTA sentence embedding methods. Unsupervised baselines include average GloVe embeddings \cite{glove}, SimCSE \cite{gao2021simcse}, DiffCSE \cite{chuang2022diffcse} and PromptBERT \cite{2022promptbert}. We choose SimCSE \cite{gao2021simcse} as the supervised baseline.
For reference, we also report two strong baselines ALIGN \cite{jia2021align} and MURAL \cite{jain-etal-2021-mural-multimodal}, which are trained specifically on MS-COCO.

% \begin{table}[t]
% \small
% \centering
% \begin{tabular}{lcc}
% \toprule
% \textbf{Model} &  \textbf{Avg. STS}  &  \textbf{CxC-STS}  \\
% \midrule
% unsup-SimCSE-BERT$_{base}$  & 76.25 & 67.5 \\
% \quad   w/ in-domain unlabeled data & 77.55 & 68.1 \\
% unsup-PromCSE-BERT$_{base}$  & \textbf{78.49} & \textbf{71.2} \\
% \quad  w/ in-domain unlabeled data & 78.47 & 71.1 \\
% \bottomrule
% \end{tabular}
% \caption{\label{tab: in-domain training}
% Test results of seven standard STS tasks (Avg. STS) and the CxC-STS task.
% }
% \end{table}

\subsubsection{Results}
% Since checkpoints of universal sentence embedding models are saved by evaluating on the labeled STS-B development set, the seven standard STS tasks in Section \ref{sec: standard sts} cannot be regarded as “out-of-domain” datasets.
% Consequently, we evaluate the zero-shot domain transfer capability by directly testing model checkpoints on the domain-shift CxC-STS dataset without further training. 
Table \ref{tab: transfer sts} demonstrates that by directly testing model checkpoints on the domain-shift CxC-STS dataset without further training, our unsupervised PromCSE remarkably boosts the performance of SimCSE by 3.7\%, with a much more significant gap than 2.2\% on standard STS tasks. Unsupervised PromCSE even outperforms state-of-the-art DiffCSE and PromptBERT by 1.1\% and 1.2\%, respectively.
Compared with supervised SimCSE, PromCSE also achieves greater improvements on the CxC-STS task than on standard STS tasks, indicating better resilience to domain shifts.
It is remarkable that our supervised PromCSE + EH could even \emph{outperform} ALIGN and MUTUAL that are trained with in-domain MS-COCO annotations, reaching new state-of-the-art results.
% These results shows that fine-tuning the whole PLM for sentence embeddings may be over-parameterized and more prone to overfit the training data, to the detriment of similar tasks in different domains. While PromCSE can effectively alleviate the domain shift problem and acquire better robustness.

\begin{table}[t]
\small
\centering
\begin{tabular}{lc}
\toprule
\multirow{2}{*}{\textbf{Model}}             & \textbf{CxC-STS}    \\
                                   & avg ± std  \\
\midrule
GloVe embeddings (avg.)$^\clubsuit$            & 55.1 ± 0.6   \\
$*$ unsup-SimCSE-BERT$_{base}$              & 67.5 ± 1.2 \\
$*$ unsup-DiffCSE-BERT$_{base}$             & 70.1 ± 1.1 \\
$*$ unsup-PromptBERT$_{base}$               & 70.0 ± 1.1 \\
$*$ unsup-PromCSE-BERT$_{base}$            & \textbf{71.2 ± 1.1} \\
\midrule
$*$ sup-SimCSE-BERT$_{base}$                & 73.0 ± 1.1 \\
$*$ sup-SimCSE-BERT$_{base}$ + EH  & 73.2 ± 1.0 \\
$*$ sup-PromCSE-BERT$_{base}$               & 73.6 ± 1.0 \\
$*$ sup-PromCSE-BERT$_{base}$ + EH & \textbf{74.0 ± 1.0} \\
\midrule
ALIGN-BERT$_{base}$$^\clubsuit$                         & 72.7 ± 0.4 \\
MURAL-BERT$_{base}$$^\clubsuit$                         & 73.9 ± 0.4 \\
\bottomrule
\end{tabular}
\caption{\label{tab: transfer sts}
Spearman’s R Bootstrap Correlation (×100) on MS-COCO 5k test set using CxC annotations. $^\clubsuit$: results from \cite{jain-etal-2021-mural-multimodal}; $*$ : results from our experiments.
}
\end{table}

\section{Ablation Studies}
\label{sec: ablation}
We investigate how different ways of choosing prompt type, prompt length and margin $m$ affect our models. We use BERT$_{base}$ model to evaluate on seven standard STS tasks and the CxC-STS task.

\begin{table}[t]
\small
\centering
\begin{tabular}{lcc}
\toprule
\textbf{Model} &  \textbf{Avg. STS}  &  \textbf{CxC-STS}  \\
\midrule
SimCSE  & 76.25 & 67.5 \\
\midrule
PromCSE  & \textbf{78.49} & \textbf{71.2} \\
\quad  layer-shared soft prompt  & 77.64 & 71.0 \\
\quad  input-layer soft prompt & 68.35 & 67.4 \\
\bottomrule
\end{tabular}
\caption{\label{tab: prompt type}
Test results of seven standard STS tasks (Avg. STS) and the CxC-STS task under different prompt types.
}
\end{table}

\begin{figure}[htp]
	\centering 
	\includegraphics[width=\columnwidth]{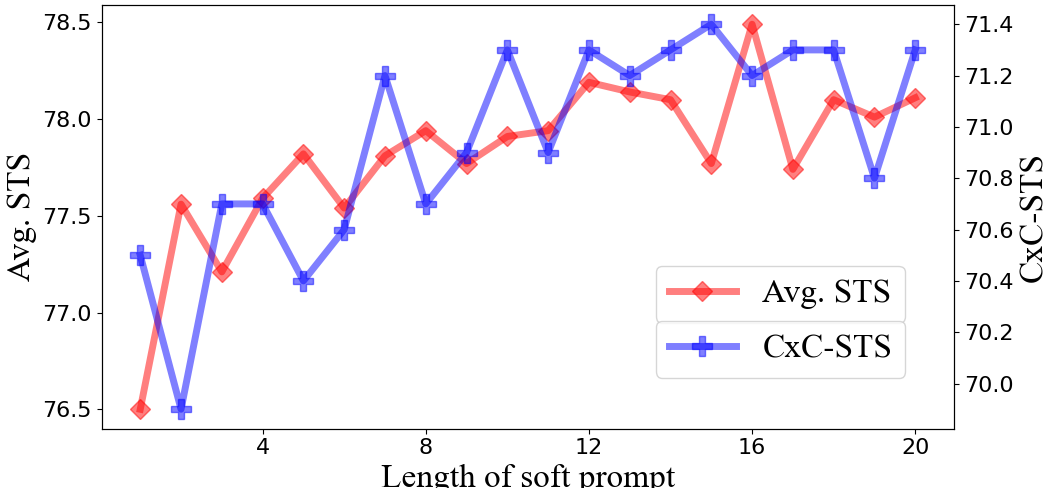}
	\caption{Test results of seven standard STS tasks (Avg. STS) and the CxC-STS task under various lengths of soft prompts.}
	\label{fig: prompt length}
\end{figure}

\begin{table}[t]
\small
\centering
\begin{tabular}{lccccc}
\toprule
$m$ & \textit{w/o} & \textit{0} & \textit{0.05} & \textit{0.1} & \textit{0.15}    \\
\textbf{Avg. STS} & 81.53 & 81.56 & 81.73 & 81.75 & 81.87 \\
\midrule
$m$ & \textit{0.2} & \textit{0.25} & \textit{0.3} & \textit{0.35} & \textit{0.4}    \\
\textbf{Avg. STS} & \textbf{81.94} & 81.91 & 81.71 & 81.48 & 81.36 \\
\bottomrule
\end{tabular}
\caption{\label{tab: margin}
The average test set results of seven standard STS tasks under different margin $m$.
}
\end{table}

\paragraph{Type of Soft Prompt}
In PromCSE, we prepend multi-layer soft prompts to PLMs instead of only the input (embedding) layer as \cite{Lester21sacle}. Table \ref{tab: prompt type} shows that only prepending soft prompts to the input layer significantly jeopardizes the performance of PromCSE on both standard STS tasks and the CxC-STS task. While making the weights of soft prompts shared across layers does not influence the effectiveness much.

\paragraph{Length of Soft Prompt}
The soft prompts in PromCSE consist of a sequence of $k$ trainable vectors. Here we regard $k$ as the length of soft prompts and investigate its effect.
Figure \ref{fig: prompt length} shows that the model performance on standard STS tasks and the CxC-STS task rises as we increase the length of soft prompts, and finally tends to stabilize when $k$ reaches around 12.
It is interesting to observe that even with $k$ set to 1, our PromCSE can still outperform SimCSE by 0.25\% on standard STS tasks and 3\% on the CxC-STS task, which indicates the effectiveness and robustness of our method. 

\paragraph{Margin $m$}
The margin $m$ in Energy-based Hinge loss (Equation (\ref{equation: eh_1})) controls the strength of the pairwise discriminative power. As shown in Table \ref{tab: margin}, the best performance is achieved when $m = 0.2$, either larger or smaller margin degrade the performance.
This matches our intuition that small $m$ may have little effect, and large $m$ may overextend the distance between negative pairs.

\section{Alignment and Uniformity Analysis}

\begin{table}[t]
\small
\centering
\begin{tabular}{lcc}
\toprule
\textbf{Model} & \textbf{Align} & \textbf{Uniform}    \\
\midrule
BERT$_{base}$ (first-last avg.)   & 0.195 & -1.304   \\
unsup-SimCSE-BERT$_{base}$   & 0.238 & \textbf{-2.337} \\
unsup-PromCSE-BERT$_{base}$  & \textbf{0.117} & -1.354 \\
\midrule
sup-SimCSE-BERT$_{base}$   & \textbf{0.241} & -3.246 \\
sup-SimCSE-BERT$_{base}$ + EH   & 0.260 & -3.349 \\
sup-PromCSE-BERT$_{base}$  & 0.325 & -3.268 \\
sup-PromCSE-BERT$_{base}$ + EH  & 0.366 & \textbf{-3.397} \\
\bottomrule
\end{tabular}
\caption{\label{tab: align}
Alignment and Uniformity measured on STS-B. The smaller numbers are better. 
% We also show the average test score of seven standard STS tasks.
}
\end{table}

% \begin{table}[t]
% \small
% \centering
% \begin{tabular}{lcc}
% \toprule
% \textbf{Dataset} & \textbf{SimCSE} & \textbf{PromCSE}    \\
% \midrule
% STS12   & 0.081 ± 0.0021 & 0.066 ± 0.0021  \\
% STS13   & 0.126 ± 0.0022 & 0.102  ± 0.0022 \\
% STS14  & 0.097 ± 0.0014 & 0.075  ± 0.0009 \\
% STS15   & 0.119 ± 0.0042 & 0.106  ± 0.0039 \\
% STS16   & 0.095 ± 0.0021 &  0.077  ±  0.0028\\
% STS-B  & 0.116 ± 0.0055 &  0.095 ± 0.0052 \\
% SICK-R  & 0.364 ± 0.0070 & 0.296 ± 0.0054 \\
% CxC-STS & 0.321 ± 0.0051 & 0.263 ± 0.0044 \\
% \bottomrule
% \end{tabular}
% \caption{\label{tab: align}
% Maximum Mean Discrepancy between the sentence embeddings of Wikipedia Corpus and those of each dataset.
% % We also show the average test score of seven standard STS tasks.
% }
% \end{table}

% Following \cite{wang2020alighment, gao2021simcse}, we utilize \textit{alignment} and \textit{uniformity} (details in Appendix \ref{sec:appendix alignment}) to measure the quality of representation space for PromCSE and SimCSE in Table \ref{tab: align}.

Alignment and uniformity are two properties proposed by \cite{wang2020alighment} to measure the quality of representations. 
Specifically, given the distribution of positive pairs $p_{pos}$ and the distribution of the whole dataset $p_{data}$, \textit{alignment} computes the expected distance between normalized embeddings of the paired sentences:
\begin{equation}
\ell_{align}\triangleq
\mathop{\mathbb{E}}\limits_{(x,x^+)\backsim p_{pos}}
\parallel f(x)-f(x^+) \parallel ^ 2
\end{equation}
While \textit{uniformity} measures how well the embeddings are uniformly distributed in the representation space:
\begin{equation}
\ell_{uniform}\triangleq
\log\mathop{\mathbb{E}}\limits_{x,y \mathop{\backsim}\limits^{i.i.d.} p_{data}}
e^{-2\parallel f(x)-f(y) \parallel ^ 2}
\end{equation}
% We utilize \textit{alignment} and \textit{uniformity} to measure the quality of representation space for PromCSE and SimCSE in Table \ref{tab: align}.
It can be seen in Table \ref{tab: align} that unsupervised PromCSE and supervised PromCSE are optimizing the representation space in two different directions. Compared with SimCSE, unsupervised PromCSE acquires better alignment, while supervised PromCSE has better uniformity. 
Besides, the Energy-based Hinge loss improves the uniformity of supervised models, which verifies its effectiveness in enhancing the pairwise discriminative power.
To directly look into the representation space of different models, we visualize the cosine similarity distribution of sentence pairs from STS-B dataset for both SimCSE and PromCSE in Appendix \ref{sec: appendix distribution}. 
It can be observed in Figure \ref{fig: similarity density} that unsupervised PromCSE preserves a lower variance while supervised PromCSE shows a more scattered distribution compared to SimCSE, corresponding to better alignment and uniformity, respectively.

\section{Conclusion}
This paper presents PromCSE, a prompt-based contrastive learning framework that improves universal sentence embeddings for resilience to domain shifts. Additionally, we theoretically show that the contrastive learning framework under NT-Xent loss is an instance of energy-based learning. To further boost the performance of supervised sentence embeddings, we propose an Energy-based Hinge loss to supplement NT-Xent loss. 
Extensive experiments on seven STS tasks and one domain shift STS task both verify the effectiveness of our method compared to current state-of-the-art supervised and unsupervised sentence embedding models.

\section*{Limitations}
In this section, we illustrate the limitations of our method. Firstly, although PromCSE outperforms SimCSE on STS tasks under both unsupervised and supervised settings, 
it cannot boost the performance of SimCSE on supervised transfer tasks, as shown in Appendix \ref{sec: appendix transfer}. 
We share a similar sentiment with \cite{ReimersG19sbert} that the primary goal of sentence embeddings is to cluster semantically similar sentences. Hence, we take STS results as the main comparison in this paper. 
Secondly, our proposed Energy-based Hinge loss is shown to be useful when hard negatives are available in supervised NLI datasets. However, how to automatically sample or generate hard negatives with unlabeled data is not discussed in this paper. We believe that designing algorithms that can automatically retrieve hard negatives will be a good direction for future work to improve the performance of unsupervised sentence embeddings.

\section*{Ethics Statement}
Since our method relies on pre-trained language models, it may run the danger of inheriting and propagating some of the models' negative biases from the data they have been pre-trained on \cite{BenderGMS21bias}. Furthermore, we do not see any other potential risks.

\bibliography{anthology,custom}
\bibliographystyle{acl_natbib}

\appendix

\section{Training Details}
\label{sec:appendix a}
We conduct experiments on 4 NVIDIA 3090Ti GPUs. The maximum sequence length is set to 32, and the temperature $\tau$ in NT-Xent loss is set to 0.05. Adam optimizer is used with a linear decay schedule.
We use grid-search of batch size $\in\{256, 512\}$, initial learning rate $\in\{$5e-3, 1e-2, 3e-2$\}$ (prompt tuning requires relative larger initial learning rate than fine-tuning) and prompt length $\in\{10, 12, 14, 16\}$. 
% We evaluate our model every 125 training steps on the development set of STS-B, and the best checkpoint is used for the final evaluation on test sets.
During the training process, we save the checkpoint with the highest score on the STS-B development set, by evaluating our model every 125 training steps. And then we use STS-B development set to find the best hyperparameters (listed in Table \ref{tab: param}).
% and we save the checkpoint with the highest score on the STS-B development set which is evaluated every 125 steps to find the best hyperparameters (listed in Table \ref{tab: param}).

\begin{table}[!h]
\small
\centering
\begin{tabular}{lcccc}
\toprule
 & Unsupervised & \multicolumn{3}{c}{Supervised}                         \\
 & BERT         & BERT & \multicolumn{2}{c}{RoBERTa} \\
 & base         & base & base         & large \\
\midrule
Batch size    & 256         & 256        & 512           & 512          \\
Learning rate & 3e-2        & 1e-2       & 1e-2         & 5e-3         \\
Prompt length & 16        & 12       & 10         & 10         \\
 \bottomrule
\end{tabular}
\caption{\label{tab: param}
The main hyperparameters for PromCSE in standard STS tasks.}
\end{table}

As for Energy-based Hinge loss, the margin $m$ is set to 0.2 according to the ablation study in Section \ref{sec: ablation}. When adding Energy-based Hinge loss to supervised SimCSE, we do not change the training configurations of the original SimCSE.

For both unsupervised and supervised PromCSE, we take the $[CLS]$ representation with an MLP layer on top of it as the sentence representation. Specially, for unsupervised PromCSE, we discard the MLP layer and only use the $[CLS]$ output during test, the same as SimCSE \cite{gao2021simcse}.

\paragraph{Prompt Initialization}
\cite{li2021prefix} find that the parameter initialization of the \textit{Soft Prompt} has a significant impact in low-data settings. 
Though our unsupervised and supervised training data both exceed 100,000, we still attempted various initialization strategies for soft prompts of PromCSE including (1) random initialization; (2) initializing with manual discrete prompt like \textit{"The meaning of the sentence"}; (3) using an LSTM to generate the sequence of \textit{Soft Prompt}; (4) first pre-training \textit{Soft Prompt} by training PromCSE using the Masked Language Modeling (MLM) objective on the training data. However, we find that different initialization strategies do not have much impact on our tasks. As a result, we randomly initialize the soft prompts using the default $init\_weights$ function provided by Huggingface’s transformers \cite{huggingface} for all the experiments.

% \section{Ablation Study of Prompt Depth}
% \label{sec:appendix b}

% \section{Alignment and Uniformity}
% \label{sec:appendix alignment}
% Alignment and uniformity are two properties proposed by \cite{wang2020alighment} to measure the quality of representations. 
% Specifically, given a distribution of positive pairs $p_{pos}$ and the distribution of the whole dataset $p_{data}$, \textit{alignment} computes the expected distance between normalized embeddings of the paired sentences:
% \begin{gather*}
% \ell_{align}\triangleq
% \mathop{\mathbb{E}}\limits_{(x,x^+)\backsim p_{pos}}
% \parallel f(x)-f(x^+) \parallel ^ 2
% \end{gather*}
% While \textit{uniformity} measures how well the embeddings are uniformly distributed in the representation space:
% \begin{gather*}
% \ell_{uniform}\triangleq
% \log\mathop{\mathbb{E}}\limits_{x,y \mathop{\backsim}\limits^{i.i.d.} p_{data}}
% e^{-2\parallel f(x)-f(y) \parallel ^ 2}
% \end{gather*}
% Smaller values of uniformity and alignment indicates better quality of the representation space.

\section{Distribution of Sentence Embeddings}
\label{sec: appendix distribution}
\begin{figure*} [t!]
	\centering
	\subfloat[\label{1a}unsup-SimCSE-BERT$_{base}$]{
		\includegraphics[width=\columnwidth]{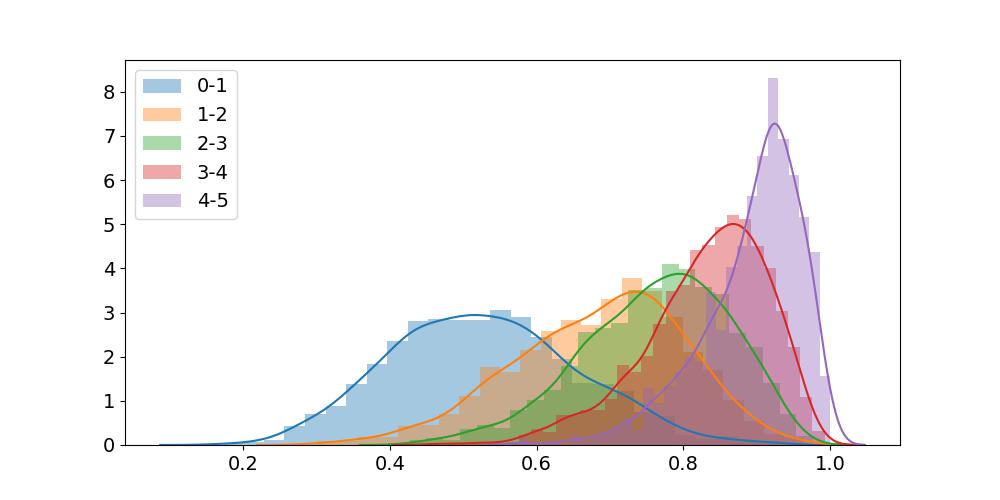}}
	\subfloat[\label{1b}unsup-PromCSE-BERT$_{base}$]{
		\includegraphics[width=\columnwidth]{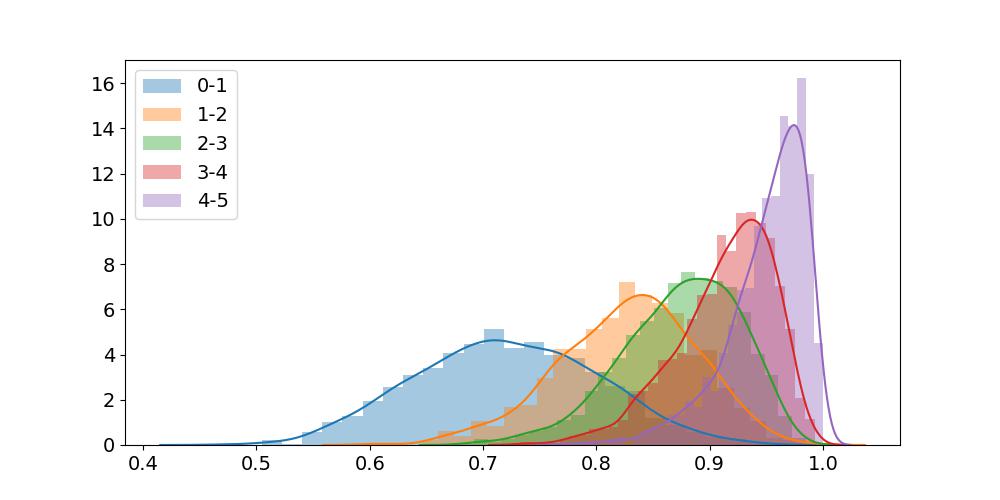}}
	\\
	\subfloat[\label{1c}sup-SimCSE-BERT$_{base}$]{
		\includegraphics[width=\columnwidth]{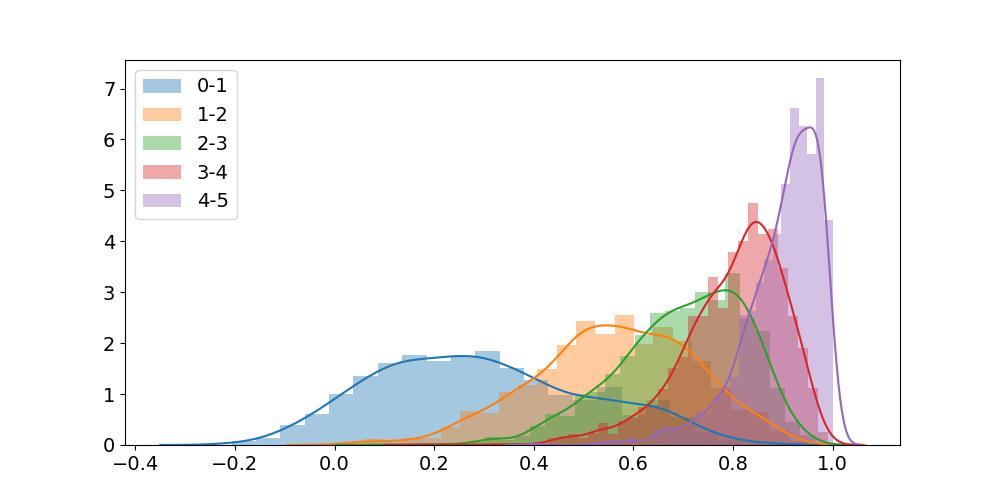}}
	\subfloat[\label{1d}sup-PromCSE-BERT$_{base}$]{
		\includegraphics[width=\columnwidth]{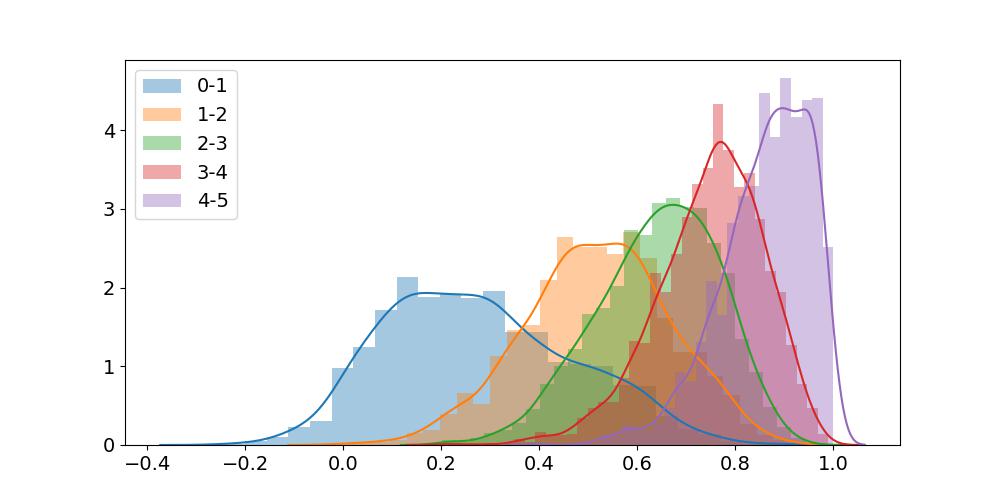}}
	\\
	\subfloat[\label{1e}sup-SimCSE-BERT$_{base}$ + EH]{
		\includegraphics[width=\columnwidth]{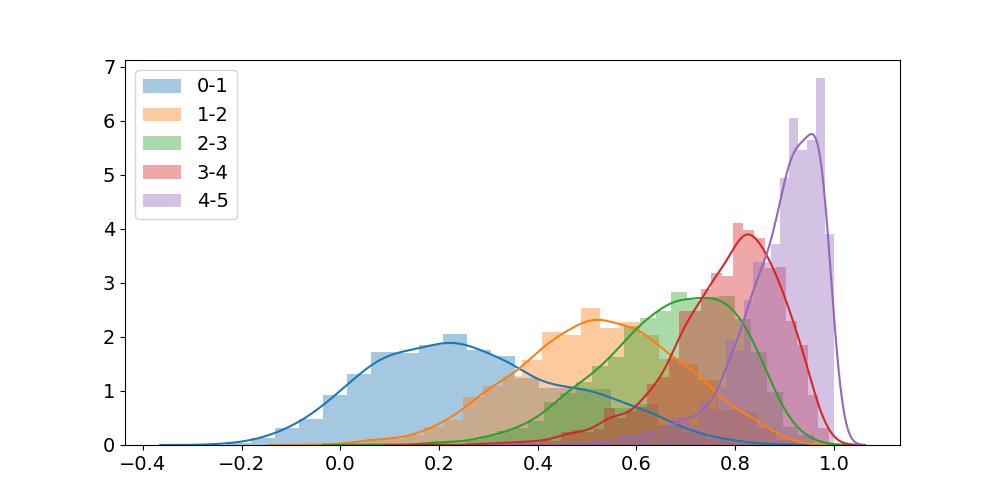}}
	\subfloat[\label{1f}sup-PromCSE-BERT$_{base}$ + EH]{
		\includegraphics[width=\columnwidth]{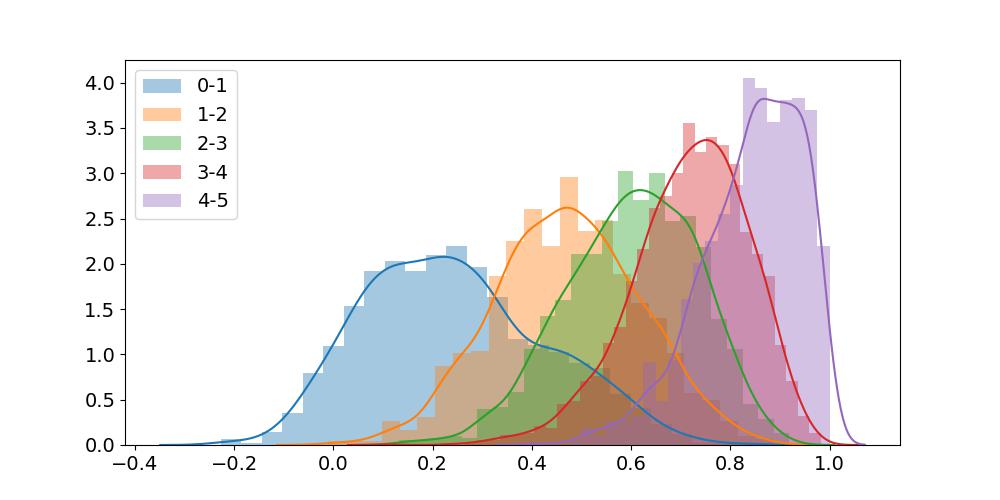}}
	\caption{Cosine Similarity Density Plots of different models between sentence pairs in STS-B. Pairs are divided into five groups based on ground truth ratings (higher means more similar). The x-axis is the model predicted cosine similarity.}
	\label{fig: similarity density}
\end{figure*}

We visualize the cosine similarity density plots of various models on the STS-Benchmark dataset in Figure \ref{fig: similarity density}. Concretely, we split the STS-B dataset into five similarity levels according to their golden labels and count all similarity scores in each sentence level. 

\section{Supervised Transfer Tasks for Sentence Embeddings}
\label{sec: appendix transfer}
Following \cite{gao2021simcse}, we evaluate our models with SentEval toolkit \cite{ConneauK18senteval} on several supervised transfer tasks, including: MR \cite{pang2005mr}, CR \cite{hu2004cr}, SUBJ \cite{pang2004subj}, MPQA \cite{wiebe2005mpqa}, SST-2 \cite{socher2013sst2} and MRPC \cite{dolan2005mrpc}. A logistic regression classifier is trained on top of (frozen) sentence embeddings produced by different methods. The evaluation results are listed in Table \ref{tab: transfer tasks} for reference.

\begin{table*}[]
\small
\centering
\begin{tabular}{lllllllll}
\toprule
\textbf{Model}                      & \textbf{MR}    & \textbf{CR}    & \textbf{SUBJ}  & \textbf{MPQA}  & \textbf{SST-2}   & \textbf{TREC}  & \textbf{MPRC}  & \textbf{Avg.}  \\
\midrule
\midrule
\multicolumn{9}{c}{\textit{Unsupervised models}} \\\midrule
GloVe embeddings (avg.)$^\clubsuit$    & 77.25 & 78.30  & 91.17 & 87.85 & 80.18 & 83.00    & 72.87 & 81.52 \\
Skip-thought$^\heartsuit$               & 76.50  & 80.10  & 93.60  & 87.10  & 82.00    & 92.20  & 73.00    & 83.50  \\
BERT$_{base}$ (first-last avg.)$^\clubsuit$ & 78.66 & 86.25 & 94.37 & 88.66 & 84.40  & 92.80  & 69.54 & 84.94 \\
BERT$_{base}$ (CLS)$^\clubsuit$             & 78.68 & 84.85 & 94.21 & 88.23 & 84.13 & 91.40  & 71.13 & 84.66 \\
IS-BERT$_{base}$$^\heartsuit$                & 81.09 & \textbf{87.18} & \textbf{94.96} & 88.75 & \textbf{85.96} & 88.64 & 74.24 & \textbf{85.83} \\
SimCSE-BERT$_{base}$$^\diamondsuit$            & \textbf{81.18} & 86.46 & 94.45 & 88.88 & 85.50  & \textbf{89.80}  & 74.43 & 85.81 \\
$*$ PromCSE-BERT$_{base}$           & 80.95 & 85.46 & 94.50  & \textbf{89.46} & 84.84 & 88.40  & \textbf{74.61} & 85.46 \\
\midrule
\midrule
\multicolumn{9}{c}{\textit{Supervised models}} \\\midrule
InferSent-GloVe$^\clubsuit$            & 81.57 & 86.54 & 92.50  & 90.38 & 84.18 & 88.20  & 75.77 & 85.59 \\
Universal Sentence Encoder$^\clubsuit$ & 80.09 & 85.19 & 93.98 & 86.70  & 86.38 & \textbf{93.20}  & 70.14 & 85.10  \\
SBERT$_{base}$$^\clubsuit$                  & \textbf{83.64} & \textbf{89.43} & 94.39 & 89.86 & \textbf{88.96} & 89.60  & \textbf{76.00}    & \textbf{87.41} \\
SimCSE-BERT$_{base}$$^\diamondsuit$            & 82.69 & 89.25 & \textbf{94.81} & 89.59 & 87.31 & 88.40  & 73.51 & 86.51 \\
$*$ SimCSE-BERT$_{base}$ + EH         &82.81       &88.82       &94.34       &89.98       &88.14       &86.20       &74.90       &86.46       \\
$*$ PromCSE-BERT$_{base}$           &81.86       &88.56       &93.78       &89.69       &86.44       &82.80       &75.36       &85.50       \\
$*$ PromCSE-BERT$_{base}$ + EH        &81.80       &89.85       &93.92       &\textbf{90.72}       &87.05       &82.60       &75.43       &85.91 \\
\bottomrule
\end{tabular}
\caption{\label{tab: transfer tasks}
Transfer task results of different sentence embedding models (measured as accuracy). $\clubsuit$: results from \cite{ReimersG19sbert}; $\heartsuit$: results from \cite{zhang-etal-2020-isbert}; $\diamondsuit$: results from \cite{gao2021simcse}; $*$ : results from our experiments; + EH: adding the Energy-based Hinge loss as shown in Equation (\ref{equation: total_loss}).
}
\end{table*}

\end{document}